\documentclass[journal]{IEEEtran}
\ifCLASSINFOpdf
   \usepackage[pdftex]{graphicx}
   \graphicspath{{../pdf/}{../jpeg/}}
   \DeclareGraphicsExtensions{.pdf,.jpeg,.png}
\else
   \usepackage[dvips]{graphicx}
   \graphicspath{{../eps/}}
   \DeclareGraphicsExtensions{.eps}
\fi

\usepackage{amsmath}
\usepackage{multirow}
\usepackage{float}

\begin{document}
%
\title{Geometric Parameter Estimations of Perovskite Solar Cells Based on Optical Simulations}
%
%
%

\author{Junhao~Wang
}

%
%

%

\markboth{Engineering Science Research Opportunities Program-Global at the National University of Singapore}{}



\maketitle

\begin{abstract}
This paper presents a non-invasive approach to estimate the layer thicknesses of perovskite solar cells. The thicknesses are predicted by a convolutional neural network that leverages the external quantum efficiency of a perovskite solar cell.  The network is trained in thickness ranges where the optical properties are constant, and these ranges set the constraints for the network’s application. Due to light sensitivity issues with opaque perovskites, the convolutional neural network showed better performance with transparent perovskites. To optimize the performance and reduce the root mean square error, we tried different sampling methods, image specifications, and Bayesian optimization for hyperparameter tuning. While sampling methods showed marginal improvement, implementing Bayesian optimization demonstrated high accuracy. Other minor optimization attempts include experimenting with input specifications and pre-processing approaches. The results confirm the feasibility, efficiency, and effectiveness of a convolution neural network for predicting perovskite solar cells’ layer thicknesses based on controlled experiments.
\end{abstract}


%
\IEEEpeerreviewmaketitle

\section{Introduction}
%
%
%
%
\IEEEPARstart{P}{erovskite} solar cells is a promising new source of solar energy production due to their high power conversion efficiency of 25.8\% comparable to currently commercialized polycrystalline silicon solar cells~\cite{min2021perovskite,green2017solar}, their stability of up to a year~\cite{grancini2017one}, and their advantageous defect tolerance~\cite{huang2017lead}.

\begin{figure}[H]
    \centering
    \includegraphics[width=2.5in]{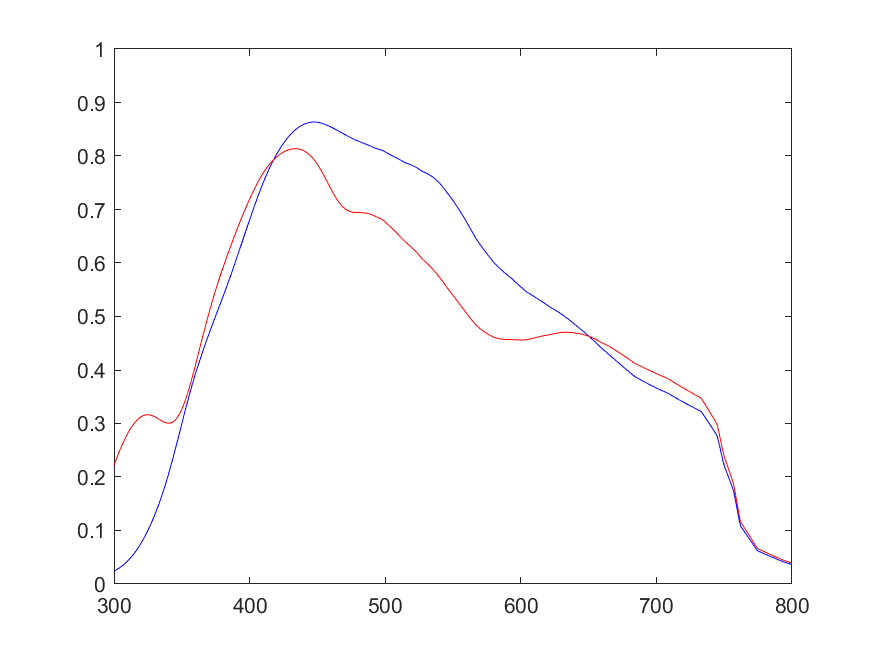}
    \caption{Example of a training image used for the transparent perovskite's CNN. Blue is forward EQE and red is reverse EQE. Note that the axes are shown here as it was used for earlier trials. The axes are removed for later trials to reduce redundant noise in the image.}
    \label{training image}
\end{figure}

We aim to determine the PSC's geometric properties based on its optical properties. The relationship between the optical property—external quantum efficiency (EQE), and the geometric property—layer thicknesses, will be studied. The EQE is the ratio between incident photons and ejected electrons at a given wavelength of light. As the EQE varies across different wavelengths, the EQE of a solar cell is often expressed over a spectrum of wavelengths, meaning that the EQE is a function of photon wavelengths (Fig.~\ref{training image}). In the perovskite cell, the external quantum efficiency curve is found to be dependent on the perovskite layer thickness~\cite{mica2020triple}. However, the relationship between the thickness of perovskite and EQE is not trivial, as different ranges of thickness have different curve shapes and peaks without an obvious trend.

We explore the estimation of layer thicknesses of perovskite solar cells using EQE functions. By using deep learning via a convolutional neural network (CNN), the approach prevents reliance on an electron microscope, which can be costly, intrusive, and unscalable. The parametric estimation also has potential applications for efficiency optimization.

Two types of perovskite are used for the CNN training and prediction in this paper; initially, the opaque perovskite, $\mathrm{Cs_{0.06}MA_{0.15}FA_{0.79}Pb(I_{0.85}Br_{0.15})_{3}}$, with the structure of an N-I-P stack layered: ITO/Sn$\mathrm{O_{2}}$/Perovskite/Spiro-OMeTAD/Au~\cite{mica2020triple}. However, as seen in Fig~\ref{optical_loss} findings of optical losses from light shining in from only one direction, a transparent perovskite is used later instead.\footnote{The optical loss could also be the problematic ranges of Spiro and Gold thickness ranges that were selected for sampling the training data. As seen in section II, we did not find evidence that indicates the ranges chosen have no optical loss.} The transparent perovskite has structure  $\mathrm{ITO/NiO/PerovHMv2/C60HM/SnO_{2}/ITO/LiF}$, with light shining from both sides of the perovskite, which resolves the optical loss problem.

\begin{figure}[H]
    \centering
    \includegraphics[width=2.5in]{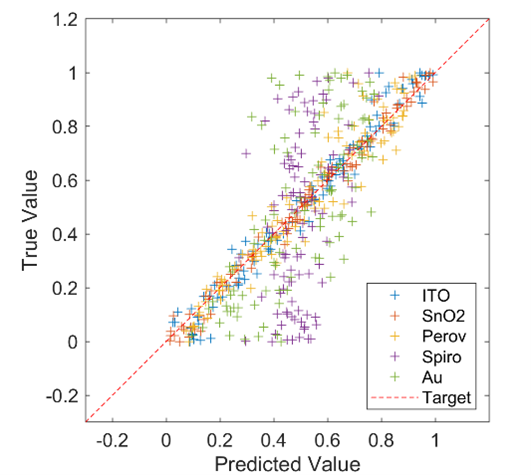}
    \caption{True vs Prediction of CNN Trained using 1000 images from Halton Sampling. Layers ITO, SnO2, and Perovskite are shown to have good predictions. Spiro-OMeTAD and Au layers, which are the farthest from the incident light, exhibit poor prediction as most light is lost before these last two layers.}
    \label{optical loss}
\end{figure}

Due to its preeminence in image recognition, CNN has various visual applications. CNN has both classifications and regression applications from graphs of ECG and EEG for pathology prediction, movement recognition, and arrhythmia prediction. These applications can be transferred to visual deep learning for perovskite cells~\cite{meng2021motor,bajpai2021automated,mathunjwa2021ecg}. Furthermore, a notable and pertinent CNN’s applications in scagnostics is found in SnagCNN, where the visual features and diagnostics scores of a scatterplot are analyzed and estimated~\cite{pham2020scagcnn}. Potentially, SnagCNN can be extended to continuous 2D graphs for visual deep learning, and the output will be adjusted to predict layer thicknesses. Another method is to make the EQE graph discrete, effectively converting it to a scatterplot before attempting to adapt SnagCNN into the estimation process. There is also research revolving around line chart understanding, which can be useful for understanding the specific graph shapes, and trends according to their “knowledge template”~\cite{sohn2021line}. SnagCNN also suggests that CNN could be suitable for the study as different layer thicknesses can have different EQE graph shapes, prevalidating the feasibility of our usage in this project.

\section{Sampling and Training Data Generation}
Sampling and generation of data are required for the CNN. We use an optical simulation known as the transfer matrix method (TMM) to calculate EQE based on layer thicknesses. To generate the data for training an inverse neural network in this relationship, the TMM is used with various sampling methods, including random, Halton, Sobol, and Latin Hypercube sampling.
\subsection{Ranges for Layer Thicknesses}
Before the sampling process, the permissible range for each layer’s thickness must be established. The range of wavelength considered is from 300 – 800 nm~\cite{mica2020triple}. A preferable thickness range is chosen such that the optical properties of the material do not significantly vary within the range. Furthermore, thicknesses should be chosen such that the perovskite produced is feasible.
\subsubsection{Opaque Pervoskite}
Firstly, the opaque perovskite solar cells’ range is established as follows.
For the ITO layer, tandem applications were conducted for thickness varying from 33 nm,  57 nm, 87 nm, and 105 nm, with the range 57-105 nm achieving optimal efficiency~\cite{bett2019semi}. A high 24.2\% efficiency of a four-terminal tandem solar cell is achieved with a 105 nm thick ITO layer~\cite{bett2019semi}. From perovskite solar cells' application in visible light communication, a 115 nm thick ITO layer is also a possibility~\cite{mica2020triple}. Furthermore, the optical properties, including the refractive index and the extinction coefficient of ITO layers that are 17 nm and 72 nm thick are similar~\cite{rii}. The thickness of ITO from 54 – 270 nm also has a constant refractive index~\cite{yan2009refractive}. However, an ITO thickness of less than 100 nm results in significantly high sheet resistance~\cite{mazur2010influence}. However, since only optical properties are considered, we will be ignoring this sheet resistance property. Based on previous research on varying the geometric properties, the ITO established range will be 54 - 350 nm~\cite{tan2022optimizing}.
The electron transport layer, SnO2, has a thickness of 25 nm commercially and 60 – 100 nm with sol-gel process~\cite{xiong2018review}. The thickness can also be 30 nm, and as thin as 10 nm in a perovskite~\cite{mica2020triple,kam2019room}. For the SnO2, the established range is 10 – 100 nm.
The perovskite layer usually varies from 60–965 nm~\cite{mica2020triple}. The thickness of the perovskite absorber can be as low as around 50 nm~\cite{della2015ultra}. To comply with the source material that demonstrates the perovskite layers’ relationship with EQE, the established range for the perovskite layer is 60–965 nm~\cite{mica2020triple}.
The Spiro-OMeTAD layer’s established range is chosen from its optimal thickness of 200-370 nm~\cite{rombach2021lessons}. The gold established range is 7-80 nm. As the gold layer gets thicker, the layer becomes more opaque. However, these optical changes become less perceivable as the range increases. It should be noted that for ranges below 80 nm, the optical losses significantly increase~\cite {yakubovsky2017optical}. Therefore, the range for Gold may need more considerations and adjustments. The ranges for each layer are specified below.
\[ \text{Subject to}
    \begin{cases} 
      \omega = [l_{\text{ITO}},l_{\text{SnO}_{2}},l_{\text{Perovskite}},l_{\text{Spiro-OMeTAD}},l_{\text{Gold}}], \\
      54 \text{ nm} \leq l_{\text{ITO}} \leq 350 \text{ nm},\\
      10 \text{ nm} \leq l_{\text{SnO}_{2}} \leq 100 \text{ nm}, \\
      60 \text{ nm} \leq l_{\text{Perovskite}} \leq 965 \text{ nm}, \\
      200 \text{ nm} \leq l_{\text{Spiro-OMeTAD}} \leq 370 \text{ nm},\\
      7 \text{ nm} \leq l_{\text{Gold}} \leq 80 \text{ nm}
   \end{cases}
\]


\subsubsection{Transparent Pervoskite}
Secondly, we will establish the transparent perovskite solar cells’ ranges. These ranges are found at the discretion of the project supervisor Hu Quee Tan.
The ITO’s range is consistent with the opaque perovskite of 54 – 350 nm.
The NiO range of 5 – 50 nm is chosen from the ranges of measured average thicknesses indicated by Wang et al., Seo et al., and Kim et al. ~\cite{wang2018high, seo2016ultra, kim2020comparison}.
The perovskite layer is the same as the opaque perovskite 60 – 965 nm.
The C60HM range of 5 – 50 nm is chosen from the studies of klipfel et al~\cite{klipfel2022c60}.
The $\text{SnO}_{2}$ range of 5 – 50 nm is restricted from the opaque perovskite and requires further investigations and adjustments.
Typically, the LiF layer is less than 1 nm thick due to its insulating properties~\cite{liu20175, huhao}. The LiF range of 50 – 300 nm also requires further adjustments and considerations. 
The ranges for each layer are specified below.

\[ \text{Subject to}
    \begin{cases} 
      \omega = [l_{\text{ITO}},l_{\text{NiO}},l_{\text{PerovHMv2}},l_{\text{C60HM}},l_{\text{SnO}_{2}},l_{\text{ITO}},l_{\text{LiF}}], \\
      54 \text{ nm} \leq l_{\text{ITO}} \leq 350 \text{ nm},\\
      5 \text{ nm} \leq l_{\text{NiO}} \leq 50 \text{ nm} ,\\
      60 \text{ nm} \leq l_{\text{PerovHMv2}} \leq 965 \text{ nm}, \\
      5 \text{ nm} \leq l_{\text{C60HM}} \leq 50 \text{ nm},\\
      5 \text{ nm} \leq l_{\text{SnO}_{2}} \leq 50 \text{ nm},\\
      50 \text{ nm} \leq l_{\text{LiF}} \leq 300 \text{ nm}
   \end{cases}
\]

\subsection{Sampling Data and Input Specifications}
\begin{figure}[H]
\centering
\includegraphics[width=3.25in]{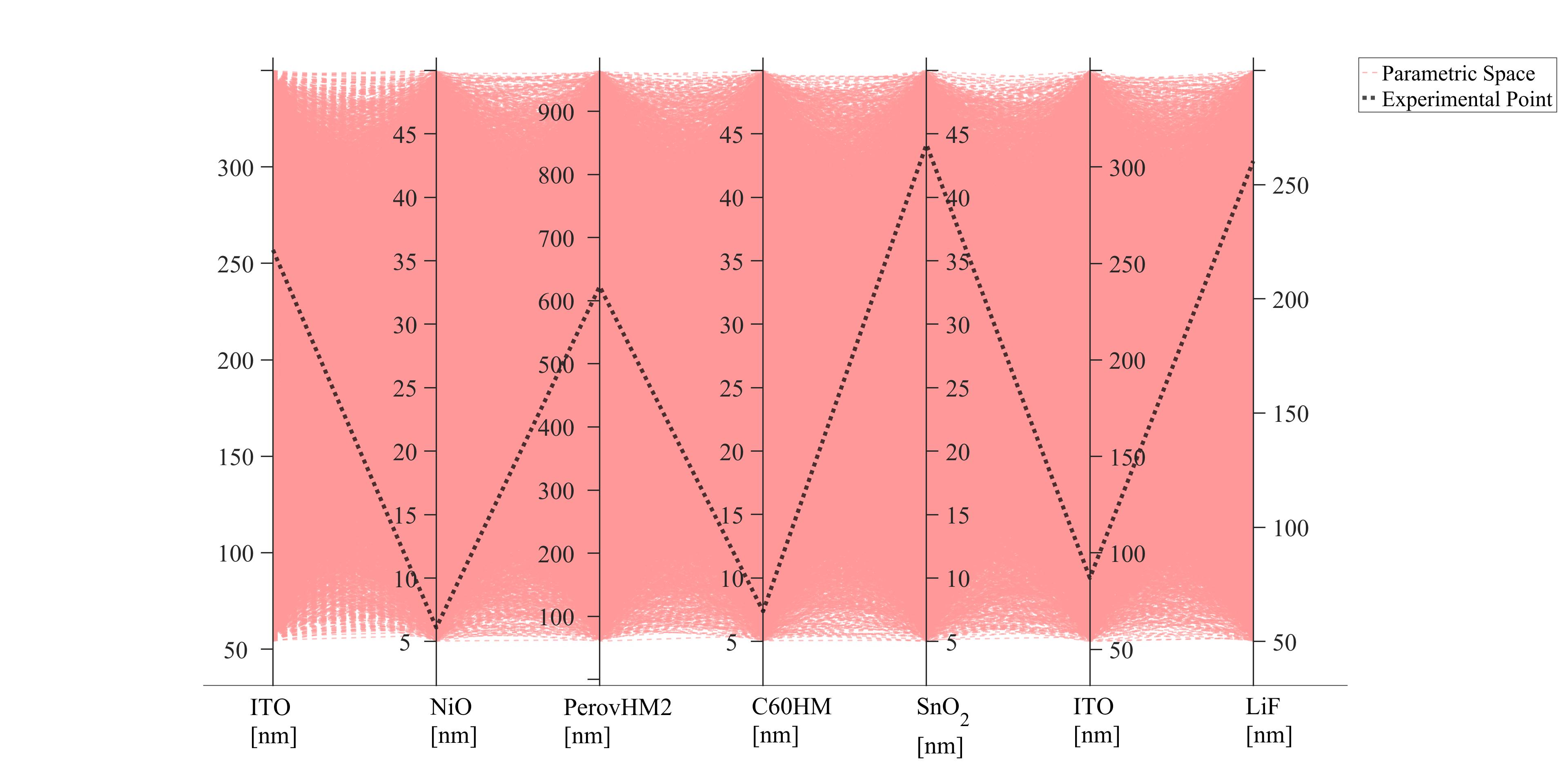}
\caption{Example of Sobol sampling to sample the required layer thicknesses that are used for generating the training dataset. Here, we sampled for transparent PSCs.}
\label{Sobol Sampling}
\end{figure}

The training, validation, and testing data is generated in the following steps. First, random, Halton, Sobol, and Latin Hypercube sampling are used to generate different combinations of layer thicknesses that are within the range defined above. This sampling is visualized in Fig.~\ref{Sobol Sampling}.

Secondly, each of the combinations is passed into the transfer matrix method function to generate the external quantum efficiency from 300 nm to 800 nm. This range was chosen due to AM1.5 intensity constraints.
The training data are pairs of layer thicknesses and external quantum efficiency.

The EQE is packaged into a MATLAB figure image in .png format. The resolution of the raw images is $875\times656$. For all figure images, the y-axis is kept between 0-1, and the x-axis is kept between 300-800 nm.

During the later trials, variations of the image specifications are experimented with. A notable variation that resulted in poorer performance is using two subplots to display the two curves for the transparent perovskite. 

\begin{figure}[H]
    \centering
    \includegraphics[width=2.5in]{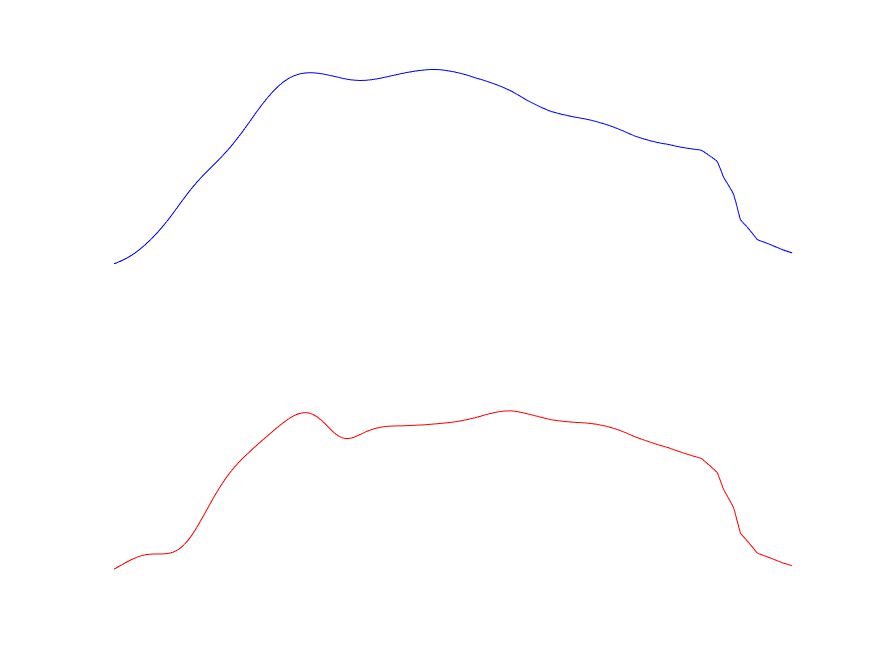}
    \caption{Example of a 2 subplot training image used for the transparent perovskite's CNN. Blue is forward EQE and red is reverse EQE. However, this plot format produced worse results.}
    \label{bad training image}
\end{figure}

12000 images are generated for each method of sampling, with 10000 as the training dataset, 1000 as the validation dataset, and 1000 as the unseen testing dataset.

The response dataset is a $12000\times7$ matrix, with each row vector representing the response layer thicknesses for the corresponding EQE figure. For this row vector, the first element is ITO thickness; the second element is NiO thickness, and so on. To normalize the data for feature scaling, we divided thicknesses by the upper bound of their corresponding layers.

\section{Training and Optimizing the Neural Network}
A convolutional neural network with appropriate architecture is constructed after attaining the training, validation, and testing datasets. The networks are created using the MATLAB Deep Learning Toolbox. The networks are trained using an MSI GS75 notebook with an i7-10750H CPU and an RTX 2060 GPU.
\subsection{First Attempt of Network Architecture}
\begin{figure}[H]
    \centering
    \includegraphics[width=3.25in]{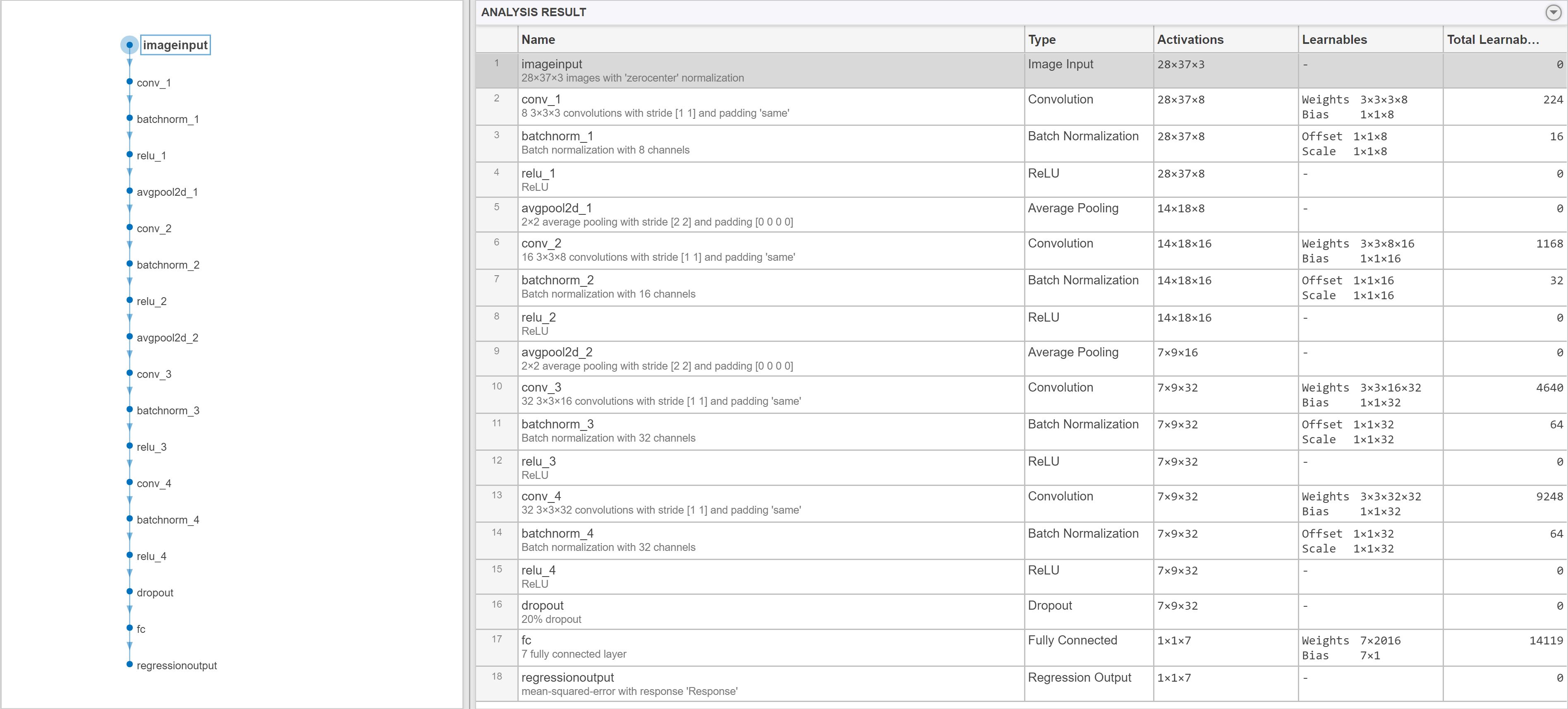}
    \caption{Our first architecture used to prevalidate that the CNN is capable of converging and predicting layer thicknesses}
    \label{simple network}
\end{figure}

At first, a relatively simplistic architecture is used for the neural network to confirm that the model is capable of converging and predicting layer thicknesses, as seen in Fig.~\ref{simple network}. This network uses stochastic gradient descent with momentum.

This network takes an input image of a resolution of $37\times28$. This resolution is chosen to match the MNIST library resolution of $28\times28$ as closely as possible. 

The network consists of four convolution layers with strides of 1 with padding. The filter sizes are all $3\times3$, and the layers consist of 8, 16, 32, and 32 filters. There are 2 average pooling layers of sizes $2\times2$, with strides of 2. Rectifier linear unit (ReLU) function and batch normalization layers are included in this network based on the MATLAB documentation. A 20\% dropout layer is also included.

\begin{figure}[H]
    \centering
    \includegraphics[width=3.25in]{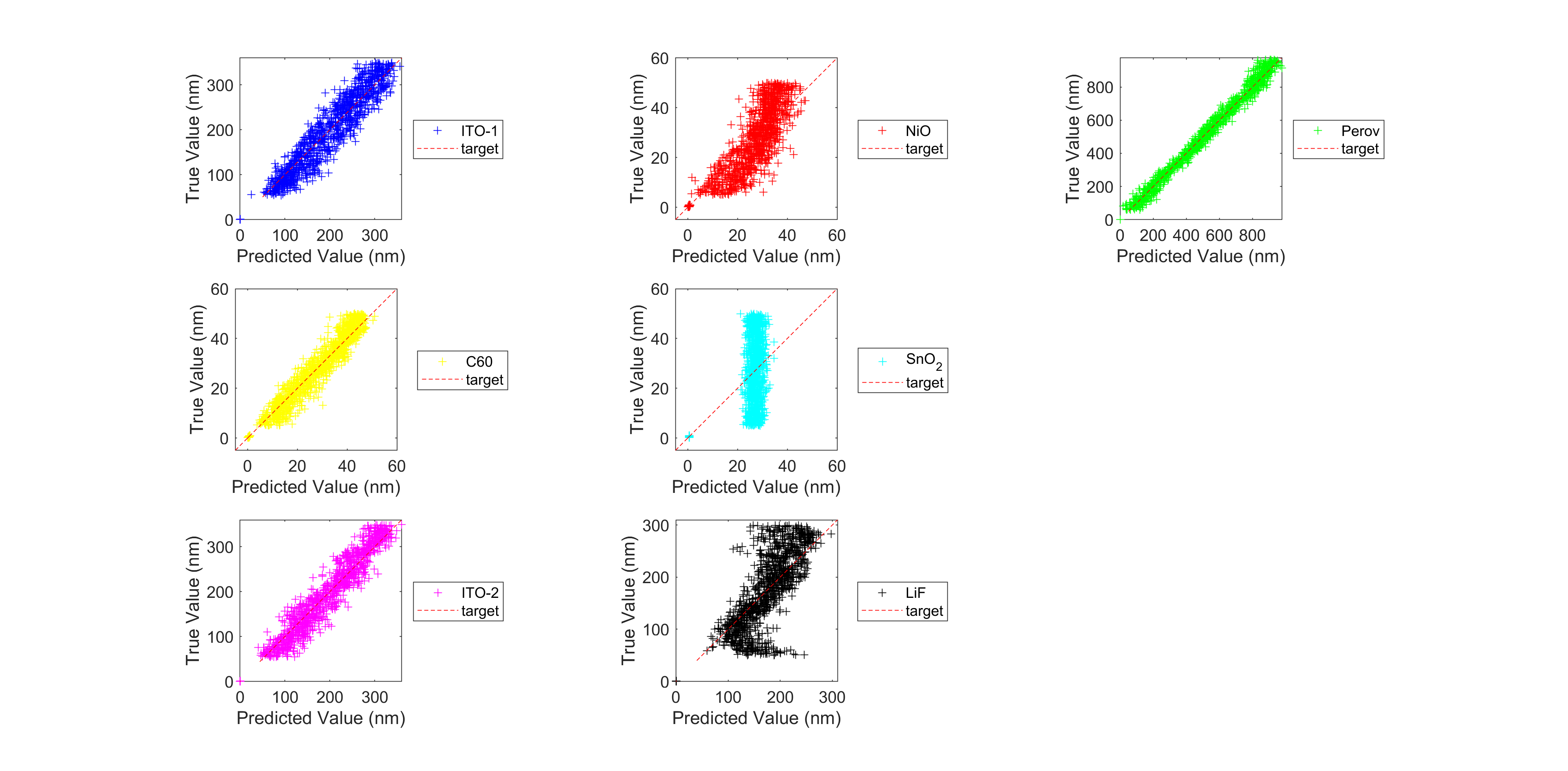}
    \caption{True vs Predicted values of layer thicknesses plots of the first network architecture without optimization. The unseen testing dataset is used. Particularly, $\text{SnO}_{2}$ and LiF layers exhibit poor accuracy.}
    \label{badwithunits}
\end{figure}

This network is capable of predicting with an overall root mean square error (RMSE) of 76.24 nm. The accuracy can be seen in Table~\ref{compare} and Fig.~\ref{badwithunits}. Notably, $\text{SnO}_{2}$ and LiF layers exhibit poor prediction.

\subsection{Bayesian Optimization}
Many experiments of Bayesian optimization are used in an attempt to optimize the CNN. The method used is mostly through trial and error and varying one hyperparameter at a time to see which variable has the largest impact on the RMSE. The hyperparameters tuned included the convolution layer's section depth, initial learning rate, momentum, mini-batch sizes, learning rate drop factors, learning rate drop period, and L2 regularization. Between experiments, we further adjusted the ranges of each hyperparameter, either narrowing them to a certain order of magnitude that has worked well or trying a different range. As a finding, we removed the dropout layer in favour of L2 regularization. After some trials, some hyperparameters are kept constant, such as the learning drop rate and overall architecture.

We ran many trials of 8 hours long experiments in conjunction with the other two optimization methods outlined below.
\subsection{Mixed Pooling Layer}
Our first architecture only used average pooling layers. Solely using average or max pooling layers could have downsides, and we have adapted our version of the mixed pooling layer based on Yu et al.~\cite{mixedpooling}. Here, we experimented with mixed usage of max and average pooling layer configurations in the network. Through trial and error, different permutations of 3 pooling layers are used to lower the RMSE further. It was found that the permutation of max, average, and average pooling layers produced the best result.

\subsection{Image and Input Layer}
We experimented with different image resolutions and input layer specifications to optimize the accuracy.

Firstly, the plot axes of images are removed to reduce the noise in the training data. This resulted in an effective decrease in RMSE.

Secondly, we experimented with a square input layer resolution of $37\times37$ and a higher input layer resolution of $100\times100$. However, there are no noticeable improvements. Particularly, the $100\times100$ resolution became too computationally heavy, and the training was too slow to complete within the 8 hours frame of the Bayesian optimization.

Thirdly, we experimented with two subplots for the transparent PSC as seen in Fig.~\ref{bad training image}. However, as mentioned, this resulted in poorer performance.

Finally, we experimented with different sampling methods for the training images. Although Sobol sampling yielded the best performances so far, we did not observe any noticeable improvements over random and Halton sampling. This could be because 12000 data points are already large enough that the sampling method no longer creates any significant differences.

\section{Results}
\begin{figure}[H]
    \centering
    \includegraphics[width=3.25in]{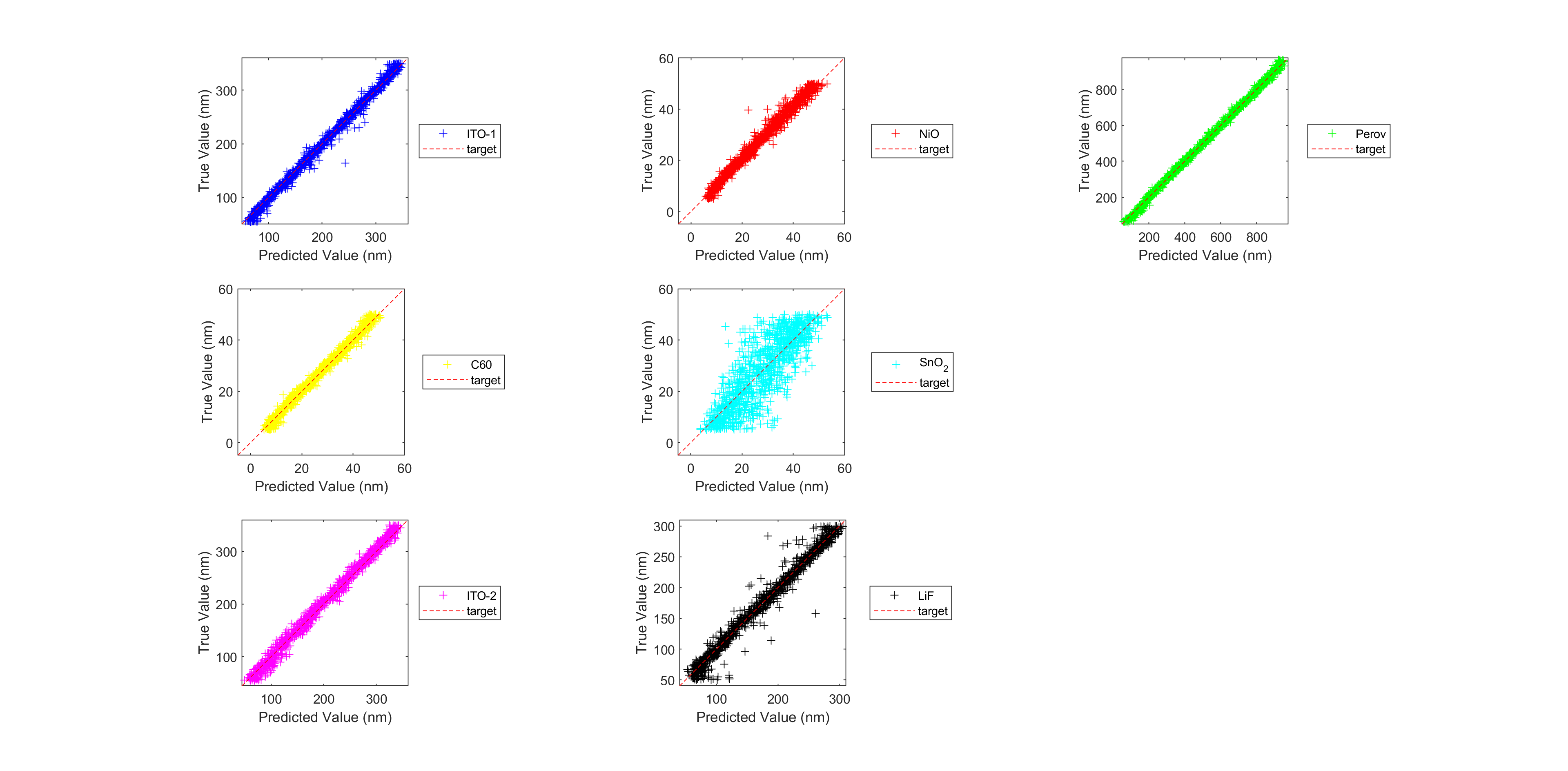}
    \caption{True vs Predicted values of layer thicknesses plots of optimized CNN. The unseen testing dataset is used. The fit of all layers improved significantly. Particularly, the LiF accuracy increased significantly, and the network can now start to predict $\text{SnO}_{2}$ as well.}
    \label{bestwithunits}
\end{figure}

\raggedbottom
The best result so far after optimizing the network can be seen in Fig.~\ref{bestwithunits} and Table~\ref{compare}. The RMSE of every layer decreased significantly. The total RMSE was reduced to 22.54 nm. Notably, NiO and C60HM layers have lower than 2 nm RMSE, and the network can now predict $\text{SnO}_{2}$ and LiF layers.

\begin{table}[H]
    \centering
    \renewcommand{\arraystretch}{1.4}
    \caption{\label{tab:temp} RMSE of CNN Before and After Optimization}
    \begin{tabular}{ |c|c|c| }
     \hline
     \multirow{2}{*}{\textbf{Layer}} & \multicolumn{2}{|c|}{\textbf{Root Mean Square Error (nm)}} \\
     \cline{2-3}
        & Before Optimization & After Optimization \\
     \hline
     ITO Top (1) & 26.74 & \textbf{7.64}\\  
     \hline
     NiO & 7.67 & \textbf{1.87}\\  
     \hline
     PerovHMv2 & 33.57 & \textbf{13.16}\\
     \hline
     C60HM & 4.24 & \textbf{1.38}\\  
     \hline
     $\text{SnO}_{2}$ & 13.27 & \textbf{7.47}\\  
     \hline
     ITO Bottom (2) & 27.32 & \textbf{8.72}\\
     \hline
     LiF & 54.52 & \textbf{11.93}\\
     \hline
     Overall & 76.24 & \textbf{22.54}\\
     \hline
    \end{tabular}
    \label{compare}
\end{table}

\begin{figure}[H]
    \centering
    \includegraphics[width=2.5in]{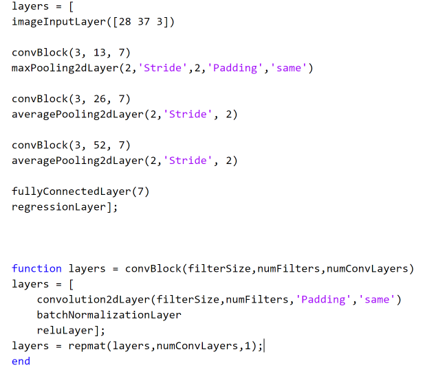}
    \caption{The final CNN architecture specified in MATLAB.}
    \label{final_architecture}
\end{figure}

Data from Sobol sampling is used to train the network. The final network uses the architecture as seen in Fig.~\ref{final_architecture}. Here, we kept the $37\times28$ input layer resolution. We used 3 blocks of convolution 2D layers, batch normalization layer, and ReLU layers; each repeated 7 times. In between the blocks, we have a max pooling layer and 2 average pooling layers.

\section{Next Steps}
There are many potential improvements and room to explore the methodologies presented in this paper. Firstly, more concepts from SnagCNN and scagnostics could be implemented into the neural network~\cite{pham2020scagcnn}. This could be done by discretizing the data points, splitting up the graph into sections, and using an ensemble network. Or, we can explore a multi-branched network of forward and reverse EQE. 

Furthermore, we should incorporate a better method to feature scale the response data during training. Currently, we used divided all the layer thickness values by the range's upper bound. Instead, a better method would be to use min-max normalization or mean normalization.

Moreover, we could experiment with mixed pooling using the formula outlined by Yu et al.~\cite{mixedpooling}.

Finally, Bayesian optimization should also be used on opaque PSC to check if optimization can alleviate the light sensitivity issue. As seen from transparent PSC, even though $\text{SnO}_{2}$ exhibited poor prediction for the first architecture, our optimization attempts significantly increased the accuracy.

\section{Conclusion}
We presented a potentially effective, accurate, and non-invasive method to determine the layer thicknesses of transparent Perovskite solar cells. Transparent PSCs are used due to the light sensitivity issue of opaque PSC's Spiro-OMeTAD and gold layer. We defined thickness ranges that have relatively constant optical properties for each layer. Then, we used 3 different sampling methods to generate sets of PSC. With each of these PSCs, we found their external quantum efficiency using the transfer matrix method. We used EQE plot figures as training data to train an inverse convolutional neural network that could predict layer thicknesses using their respective EQE plots. To train the network, we first created a relatively simple neural network and optimized it using Bayesian optimization. With multiple optimization attempts, we were able to achieve a relatively high accuracy of 22.54 nm RMSE for the norm of all 7 layers, with NiO and C60HM layers having lower than 2 nm RMSE and the perovskite layer achieving low root mean square percent errors. Overall, this method could greatly reduce the cost and complexity of measuring the geometric properties of PSC and have applications for future research.


%

\section*{Acknowledgment}
The authors would like to thank Professor Erik Birgersson, Hu Quee Tan, and Xinhai Zhao for their support and mentorship, as well as the Department of Engineering Science, the University of Toronto, SERIS, and the National University of Singapore for providing the opportunity to work on this project.

\ifCLASSOPTIONcaptionsoff
  \newpage
\fi



%



\bibliographystyle{IEEEtran}
\bibliography{bibliography.bib}

\begin{thebibliography}{10}
\providecommand{\url}[1]{#1}
\csname url@samestyle\endcsname
\providecommand{\newblock}{\relax}
\providecommand{\bibinfo}[2]{#2}
\providecommand{\BIBentrySTDinterwordspacing}{\spaceskip=0pt\relax}
\providecommand{\BIBentryALTinterwordstretchfactor}{4}
\providecommand{\BIBentryALTinterwordspacing}{\spaceskip=\fontdimen2\font plus
\BIBentryALTinterwordstretchfactor\fontdimen3\font minus
  \fontdimen4\font\relax}
\providecommand{\BIBforeignlanguage}[2]{{%
\expandafter\ifx\csname l@#1\endcsname\relax
\typeout{** WARNING: IEEEtran.bst: No hyphenation pattern has been}%
\typeout{** loaded for the language `#1'. Using the pattern for}%
\typeout{** the default language instead.}%
\else
\language=\csname l@#1\endcsname
\fi
#2}}
\providecommand{\BIBdecl}{\relax}
\BIBdecl

\bibitem{min2021perovskite}
H.~Min, D.~Y. Lee, J.~Kim, G.~Kim, K.~S. Lee, J.~Kim, M.~J. Paik, Y.~K. Kim,
  K.~S. Kim, M.~G. Kim \emph{et~al.}, ``Perovskite solar cells with atomically
  coherent interlayers on sno2 electrodes,'' \emph{Nature}, vol. 598, no. 7881,
  pp. 444--450, 2021.

\bibitem{green2017solar}
M.~A. Green, Y.~Hishikawa, W.~Warta, E.~D. Dunlop, D.~H. Levi, J.~Hohl-Ebinger,
  and A.~W. Ho-Baillie, ``Solar cell efficiency tables (version 50),''
  \emph{Progress in photovoltaics: research and applications}, vol.~25, no.~7,
  pp. 668--676, 2017.

\bibitem{grancini2017one}
G.~Grancini, C.~Rold{\'a}n-Carmona, I.~Zimmermann, E.~Mosconi, X.~Lee,
  D.~Martineau, S.~Narbey, F.~Oswald, F.~De~Angelis, M.~Graetzel \emph{et~al.},
  ``One-year stable perovskite solar cells by 2d/3d interface engineering,''
  \emph{Nature communications}, vol.~8, no.~1, pp. 1--8, 2017.

\bibitem{huang2017lead}
H.~Huang, M.~I. Bodnarchuk, S.~V. Kershaw, M.~V. Kovalenko, and A.~L. Rogach,
  ``Lead halide perovskite nanocrystals in the research spotlight: stability
  and defect tolerance,'' \emph{ACS energy letters}, vol.~2, no.~9, pp.
  2071--2083, 2017.

\bibitem{mica2020triple}
N.~A. Mica, R.~Bian, P.~Manousiadis, L.~K. Jagadamma, I.~Tavakkolnia, H.~Haas,
  G.~A. Turnbull, and I.~D. Samuel, ``Triple-cation perovskite solar cells for
  visible light communications,'' \emph{Photonics Research}, vol.~8, no.~8, pp.
  A16--A24, 2020.

\bibitem{meng2021motor}
X.~Meng, S.~Qiu, S.~Wan, K.~Cheng, and L.~Cui, ``A motor imagery eeg signal
  classification algorithm based on recurrence plot convolution neural
  network,'' \emph{Pattern Recognition Letters}, vol. 146, pp. 134--141, 2021.

\bibitem{bajpai2021automated}
R.~Bajpai, R.~Yuvaraj, and A.~A. Prince, ``Automated eeg pathology detection
  based on different convolutional neural network models: Deep learning
  approach,'' \emph{Computers in Biology and Medicine}, vol. 133, p. 104434,
  2021.

\bibitem{mathunjwa2021ecg}
B.~M. Mathunjwa, Y.-T. Lin, C.-H. Lin, M.~F. Abbod, and J.-S. Shieh, ``Ecg
  arrhythmia classification by using a recurrence plot and convolutional neural
  network,'' \emph{Biomedical Signal Processing and Control}, vol.~64, p.
  102262, 2021.

\bibitem{pham2020scagcnn}
V.~Pham, N.~V. Nguyen, and T.~Dang, ``Scagcnn: Estimating visual
  characterizations of 2d scatterplots via convolution neural network,'' in
  \emph{Proceedings of the 11th International Conference on Advances in
  Information Technology}, 2020, pp. 1--9.

\bibitem{sohn2021line}
C.~Sohn, H.~Choi, K.~Kim, J.~Park, and J.~Noh, ``Line chart understanding with
  convolutional neural network,'' \emph{Electronics}, vol.~10, no.~6, p. 749,
  2021.

\bibitem{bett2019semi}
A.~J. Bett, K.~M. Winkler, M.~Bivour, L.~Cojocaru, O.~S. Kabakli, P.~S.
  Schulze, G.~Siefer, L.~Tutsch, M.~Hermle, S.~W. Glunz \emph{et~al.},
  ``Semi-transparent perovskite solar cells with ito directly sputtered on
  spiro-ometad for tandem applications,'' \emph{ACS applied materials \&
  interfaces}, vol.~11, no.~49, pp. 45\,796--45\,804, 2019.

\bibitem{rii}
M.~N. Polyanskiy, ``Refractive index database,''
  \url{https://refractiveindex.info}, accessed on 2022-06-10.

\bibitem{yan2009refractive}
X.~Yan, F.~W. Mont, D.~J. Poxson, M.~F. Schubert, J.~K. Kim, J.~Cho, and E.~F.
  Schubert, ``Refractive-index-matched indium--tin-oxide electrodes for liquid
  crystal displays,'' \emph{Japanese Journal of Applied Physics}, vol.~48, no.
  12R, p. 120203, 2009.

\bibitem{mazur2010influence}
M.~Mazur, D.~Kaczmarek, J.~Domaradzki, D.~Wojcieszak, S.~Song, and F.~Placido,
  ``Influence of thickness on transparency and sheet resistance of ito thin
  films,'' in \emph{The Eighth International Conference on Advanced
  Semiconductor Devices and Microsystems}.\hskip 1em plus 0.5em minus
  0.4em\relax IEEE, 2010, pp. 65--68.

\bibitem{tan2022optimizing}
H.~Q. Tan, X.~Zhao, A.~Jiao, E.~Birgersson, and H.~Xue, ``Optimizing bifacial
  all-perovskite tandem solar cell: How to balance light absorption and
  recombination,'' \emph{Solar Energy}, vol. 231, pp. 1092--1106, 2022.

\bibitem{xiong2018review}
L.~Xiong, Y.~Guo, J.~Wen, H.~Liu, G.~Yang, P.~Qin, and G.~Fang, ``Review on the
  application of sno2 in perovskite solar cells,'' \emph{Advanced Functional
  Materials}, vol.~28, no.~35, p. 1802757, 2018.

\bibitem{kam2019room}
M.~Kam, Q.~Zhang, D.~Zhang, and Z.~Fan, ``Room-temperature sputtered sno2 as
  robust electron transport layer for air-stable and efficient perovskite solar
  cells on rigid and flexible substrates,'' \emph{Scientific reports}, vol.~9,
  no.~1, pp. 1--10, 2019.

\bibitem{della2015ultra}
E.~Della~Gaspera, Y.~Peng, Q.~Hou, L.~Spiccia, U.~Bach, J.~J. Jasieniak, and
  Y.-B. Cheng, ``Ultra-thin high efficiency semitransparent perovskite solar
  cells,'' \emph{Nano Energy}, vol.~13, pp. 249--257, 2015.

\bibitem{rombach2021lessons}
F.~M. Rombach, S.~A. Haque, and T.~J. Macdonald, ``Lessons learned from
  spiro-ometad and ptaa in perovskite solar cells,'' \emph{Energy \&
  Environmental Science}, 2021.

\bibitem{yakubovsky2017optical}
D.~I. Yakubovsky, A.~V. Arsenin, Y.~V. Stebunov, D.~Y. Fedyanin, and V.~S.
  Volkov, ``Optical constants and structural properties of thin gold films,''
  \emph{Optics express}, vol.~25, no.~21, pp. 25\,574--25\,587, 2017.

\bibitem{wang2018high}
T.~Wang, D.~Ding, X.~Wang, R.~Zeng, H.~Liu, and W.~Shen, ``High-performance
  inverted perovskite solar cells with mesoporous nio x hole transport layer by
  electrochemical deposition,'' \emph{ACS omega}, vol.~3, no.~12, pp.
  18\,434--18\,443, 2018.

\bibitem{seo2016ultra}
S.~Seo, I.~J. Park, M.~Kim, S.~Lee, C.~Bae, H.~S. Jung, N.-G. Park, J.~Y. Kim,
  and H.~Shin, ``An ultra-thin, un-doped nio hole transporting layer of highly
  efficient (16.4\%) organic--inorganic hybrid perovskite solar cells,''
  \emph{Nanoscale}, vol.~8, no.~22, pp. 11\,403--11\,412, 2016.

\bibitem{kim2020comparison}
S.-K. Kim, H.-J. Seok, D.-H. Kim, D.-H. Choi, S.-J. Nam, S.-C. Kim, and H.-K.
  Kim, ``Comparison of nio x thin film deposited by spin-coating or thermal
  evaporation for application as a hole transport layer of perovskite solar
  cells,'' \emph{RSC advances}, vol.~10, no.~71, pp. 43\,847--43\,852, 2020.

\bibitem{klipfel2022c60}
N.~Klipfel, A.~O. Alvarez, H.~Kanda, A.~A. Sutanto, C.~Igci, C.~Roldan-Carmona,
  C.~Momblona, F.~Fabregat-Santiago, and M.~K. Nazeeruddin, ``C60 thin films in
  perovskite solar cells: Efficient or limiting charge transport layer?''
  \emph{ACS Applied Energy Materials}, vol.~5, no.~2, pp. 1646--1655, 2022.

\bibitem{liu20175}
X.~Liu, L.~J. Guo, and Y.~Zheng, ``5-nm lif as an efficient cathode buffer
  layer in polymer solar cells through simply introducing a c60 interlayer,''
  \emph{Nanoscale research letters}, vol.~12, no.~1, pp. 1--7, 2017.

\bibitem{huhao}
H.~Hu, K.~Wong, T.~Kollek, F.~Hanusch, S.~Polarz, P.~Docampo, and
  L.~Schmidt-Mende, ``Highly efficient reproducible perovskite solar cells
  prepared by low-temperature processing,'' \emph{Molecules}, vol.~21, p. 542,
  04 2016.

\bibitem{mixedpooling}
D.~Yu, H.~Wang, P.~Chen, and Z.~Wei, ``Mixed pooling for convolutional neural
  networks,'' 10 2014, pp. 364--375.

\end{thebibliography}

%




\end{document}